\documentclass{article}
\usepackage{spconf,amsmath,graphicx}

\usepackage{enumitem}
\setlist{nosep, leftmargin=14pt}

\usepackage[noadjust]{cite}
\usepackage{amssymb,amsfonts}
\usepackage{algorithmic}
\usepackage{textcomp}
\usepackage{xcolor}
\usepackage{booktabs} 
\usepackage{subcaption}
\usepackage{multirow}

\usepackage{pifont}
\usepackage[hidelinks]{hyperref}

\title{Visual Fixation-Based Retinal Prosthetic Simulation}

\name{%
\begin{tabular}{ccccc}
Yuli Wu$^{\,1}$ & 
Do Dinh Tan Nguyen$^{\,1}$& 
Henning Konermann$^{\,1}$& \\
Rüveyda Yilmaz$^{\,1}$& 
Peter Walter$^{\,2}$ & 
Johannes Stegmaier$^{\,1}$\vspace{-0.3cm}
\end{tabular}}

\address{\normalsize $^{1}$ Institute of Imaging and Computer Vision, RWTH Aachen University, Aachen, Germany \\
\normalsize$^{2}$ Department of Ophthalmology, RWTH Aachen University, Aachen, Germany \\
\normalsize E-mail: \texttt{\small\{yuli.wu, johannes.stegmaier\}@lfb.rwth-aachen.de}}

\begin{document}

\maketitle
\begin{abstract}
This study proposes a retinal prosthetic simulation framework driven by visual fixations, inspired by the saccade mechanism, and assesses performance improvements through end-to-end optimization in a classification task.
Salient patches are predicted from input images using the self-attention map of a vision transformer to mimic visual fixations. These patches are then encoded by a trainable U-Net and simulated using the pulse2percept framework to predict visual percepts.
By incorporating a learnable encoder, we aim to optimize the visual information transmitted to the retinal implant, addressing both the limited resolution of the electrode array and the distortion between the input stimuli and resulting phosphenes.
The predicted percepts are evaluated using the self-supervised DINOv2 foundation model, with an optional learnable linear layer for classification accuracy.
On a subset of the ImageNet validation set, the fixation-based framework achieves a classification accuracy of 87.72\%, using computational parameters based on a real subject's physiological data, significantly outperforming the downsampling-based accuracy of 40.59\% and approaching the healthy upper bound of 92.76\%.
Our approach shows promising potential for producing more semantically understandable percepts with the limited resolution available in retinal prosthetics.

\end{abstract}


\section{Introduction}

Retinal implants hold significant promise for restoring visual perception in individuals affected by conditions such as retinitis pigmentosa and age-related macular degeneration. However, existing implants often face limitations in the quality of visual phosphenes (perceived light flashes) \cite{Ayton_2020,michael_beyeler-proc-scipy-2017} due to the restricted number of electrodes, such as the 60 electrodes in the Argus II device \cite{luo2016argus}. To overcome these challenges, deep learning-based stimulus encoders have been developed using methods, \textit{e.g.}, autoencoders \cite{Granley2022hybrid}, end-to-end optimization \cite{Relic2022deep,RuytervanSteveninck2022,wu2023embc}, human-in-the-loop optimization \cite{Granley2023human}, and invertible neural networks \cite{wu2024optimizing}. 
These encoders aim to reduce distortion in the predicted percepts by training neural networks to invert models of phosphene formation and optimize the input stimulus (the electrical signals delivered to the electrode array of a retinal implant) \textit{w.r.t.} pixelwise or perceptual losses.
When dealing with high-resolution input images, current approaches (\textit{e.g.}, \cite{Granley2022hybrid,wu2024optimizing}) often downscale them to fit the limited size of the electrode array, which results in significant information loss and does not accurately reflect how patients use visual prostheses to identify objects in practice. Alternatively, scanning techniques are commonly employed to help identify the semantics of large objects, but this requires subjects to move their heads \cite{EricksonDavis2021}. These limitations motivate us to mimic the inherent saccade mechanism of human eyes to address the issue in the retinal prosthetic simulation.

\begin{figure}[b]
  \centering
  \begin{subfigure}{0.11\textwidth}
    \includegraphics[width=\textwidth]{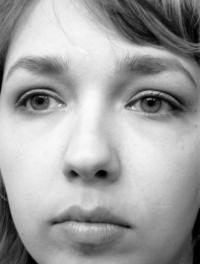}
    \caption{}
    \label{fig:saccade-1}
  \end{subfigure}\hspace{0.02\textwidth}
  \begin{subfigure}{0.11\textwidth}
    \includegraphics[width=\textwidth]{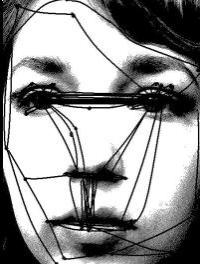}
    \caption{}
    \label{fig:saccade-2}
  \end{subfigure}\hspace{0.02\textwidth}
  \begin{subfigure}{0.11\textwidth}
    \includegraphics[width=\textwidth]{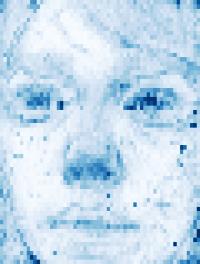}
    \caption{}
    \label{fig:saccade-3}
  \end{subfigure}
  \vspace{-0.6em}
  \caption{(a) Original image. (b) Trace of saccades of the human eye viewing a still image (\cite{wiki:saccade} under CC BY-SA 2.0). (c) Self-attention from a Vision Transformer in DINOv2 \cite{oquab2024dinov}.}
  \label{fig:saccade}
\end{figure}

Saccades are rapid, abrupt eye movements and play a crucial role in visual perception by allowing us to shift our gaze quickly between different points of interest \cite{martinez2004role}. Between these movements, the eyes pause in fixations, where visual information is gathered and processed, as illustrated in Fig.~\ref{fig:saccade}. Various approaches are proposed to predict the visual fixations with deep learning techniques, \textit{e.g.}, by incorporating attention mechanisms \cite{Kuemmerer2022} and by leveraging recurrent neural networks \cite{cornia2018predicting}.
In this work, we propose that optimizing prosthetic vision should account for the natural occurrence of saccadic movements to improve the identification of large objects, rather than focusing solely on the perception of individual static images.  The fixation-based approach outperforms downsampling-based methods in terms of classification accuracy using a foundation model, offering an alternative framework for retinal prosthetic simulations. 

\begin{figure*}[t]
    \centering
    \includegraphics[width=.97\linewidth]{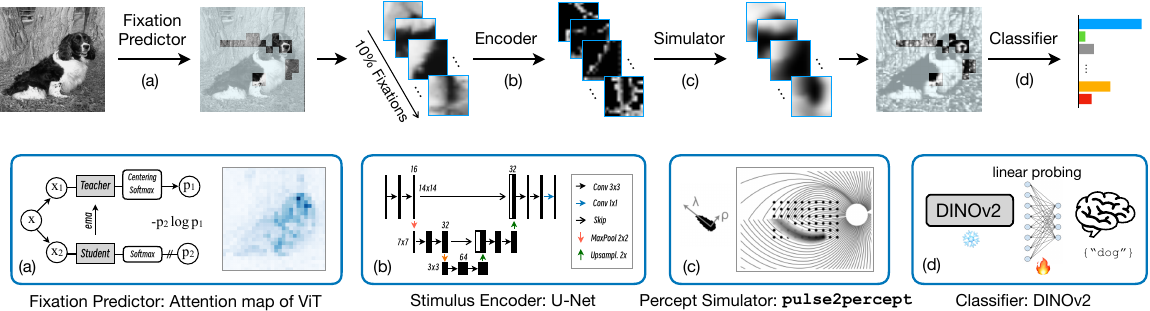}
    \caption{Overview of the visual fixation-based retinal prosthetic simulation pipeline. Salient patches are extracted using a fixation predictor informed by the attention scores from a Vision Transformer (ViT) \cite{dosovitskiy2021an}, as implemented in the self-supervised DINOv2 pre-trained foundation model \cite{caron2021emerging,oquab2024dinov}. These fixation patches are then encoded by a U-Net \cite{ronneberger2015u} to optimize the stimulus, which is trained for classification tasks on simulated percepts generated by the \texttt{pulse2percept} framework \cite{michael_beyeler-proc-scipy-2017}. The performance is evaluated using the frozen DINOv2 backbone, with the option of applying learnable linear probing.}
    \label{fig:overview}
\end{figure*}

\section{Materials and Methods}
\subsection{Dataset}
As a proof of concept, we utilize Imagenette\footnote{\url{https://github.com/fastai/imagenette}}, a subset of the ImageNet dataset \cite{deng2009imagenet}, for simulating the visual fixation-based retinal prosthesis. 
Imagenette contains 10 easily distinguishable classes, with 9,469 training images and 3,925 validation images. 
Given the minimal class imbalance, we report classification accuracy as the end-to-end evaluation metric.

\begin{figure}[b]
\begin{minipage}[b]{1\linewidth}
\begin{minipage}[b]{0.4\textwidth}
    \centering
    \includegraphics[width=\textwidth]{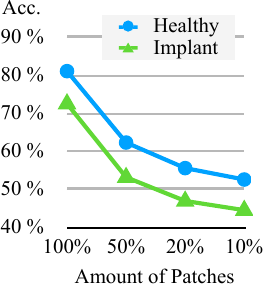}
    \captionof{figure}{Classification accuracy ($x$) \textit{w.r.t.} the ratio of the most salient fixation patches ($y$).}
    \label{fig:fixation}
\end{minipage}
\hfill
\begin{minipage}[b]{0.54\textwidth}
\centering
    \includegraphics[width=\textwidth]{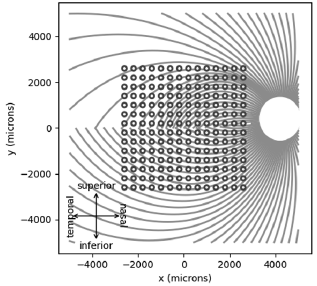}
    \captionof{figure}{A virtual retinal implant with 196 electrodes (14$\times$14) on the retinal axon map \cite{michael_beyeler-proc-scipy-2017,beyeler2019model}.}
    \label{fig:implant}
\end{minipage}
\end{minipage}
\end{figure}

\subsection{Fixation Predictor}\label{sec:salient}
We leverage the self-attention \cite{vaswani2017attention} of the \texttt{[CLS]} token on the heads of the last layer from an ImageNet pre-trained Vision Transformer (ViT \cite{dosovitskiy2021an}) and regard the attention score as the saliency of the fixation in the saccade mechanism of the human eye \cite{martinez2004role}. As illustrated in Fig.~\ref{fig:saccade}, the self-attention mask highlights regions similar to those traced by saccades.
Out of the 256 patches (16$\times$16) from each input image (resized to 224$\times$224 pixels), the top 10\% (\textit{i.e.}, \texttt{int(25.6)=25}) most salient fixation patches are preserved based on the sum within each patch of the multi-head attention \cite{vaswani2017attention}, as shown in Fig.~\ref{fig:overview}. 

The percentage of remaining salient patches is determined based on the classification accuracy of a pre-trained ViT using masked images, similar to the approach used in masked autoencoders (MAE) \cite{he2022masked}. The non-salient patches are masked with a value of 0, while their positions are preserved to maintain the validity of the positional encoding. We employ the ImageNet-1k dataset \cite{deng2009imagenet} with the DINOv2 ViT-S/14 distilled model \cite{caron2021emerging, oquab2024dinov} to investigate the relationship between the percentage of salient fixations and classification accuracy. 
As shown in Fig.~\ref{fig:fixation}, although the classification accuracy declines as fewer patches are available, the model still achieves a reasonable 52.59\% accuracy for 1,000 classes with only 10\% of the salient patches, especially when compared to the approximately 0.1\% accuracy expected from random guessing. 
For the simpler Imagenette subset, an accuracy of 92.76\% is achieved with just 10\% of the patches, providing a strong upper bound for comparison in the following experiments.

\subsection{Retinal Prosthetic Encoder}\label{sec:encoder}
To optimize the electrical stimulus delivered to the electrode array of a retinal implant, we utilize a neural network-based encoder positioned at the interface between the camera and the electrode array, following the approach outlined in \cite{wu2023embc}. This learnable encoder receives salient patches as input and produces an optimized representation, which is subsequently passed to the Axon Map Model \cite{beyeler2019model} implemented in the \texttt{pulse2percept} framework \cite{michael_beyeler-proc-scipy-2017}.
We use a shallow U-Net architecture \cite{ronneberger2015u} as the encoder, consisting of three layers with two downsampling and upsampling blocks, to efficiently process and refine the input patches. 
The retinal prosthetic encoder is trained in an end-to-end manner, with the classification loss cross-entropy driving the optimization process, as depicted in Fig.~\ref{fig:overview}. 
By incorporating this learnable encoder, we aim to improve the quality of the visual information delivered to the retinal implant given the limited resolution of the electrode array and the distortion from the electrical signals to the percepts, ultimately enhancing the perception experience for the patients with impaired vision.

\subsection{Retinal Prosthetic Simulator}\label{sec:simulator}

 The visual phosphenes elicited by the axons, somas, or dendrites of retinal ganglion cells take on distinct shapes \cite{greenberg1999computational}. These shapes have been validated through drawings provided by retinal implant recipients and modeled using the Axon Map Model \cite{beyeler2019model} in \texttt{pulse2percept} \cite{michael_beyeler-proc-scipy-2017}.
In our \textit{in silico} experiments, we simulate a virtual square retinal implant with a $14\times 14$ electrode grid. This configuration is feasible for production and compatible with pre-trained ViT weights (Fig.\ref{fig:implant}). To predict the resulting visual percept, we employ the physiologically validated Axon Map Model \cite{beyeler2019model} as the computational framework. The percept intensity is computed using exponential decay functions based on two factors: the distance from the stimulus center and the distance along retinal nerve fiber bundle trajectories to the elongated soma. These decays are governed by two constants: the radial decay rate $\rho$ and the axonal decay rate $\lambda$ (Fig. \ref{fig:overview}c).
For this study, we utilize two parameter sets: an idealized case with minimal distortion, and a more realistic but challenging case derived from subject-specific data in \cite{beyeler2019model}.

\subsection{Feature Extractor and Classifier}\label{sec:dinov2}

DINOv2 is a pretrained, self-supervised model designed to efficiently generate consistent visual representations for unseen datasets, without requiring fine-tuning \cite{oquab2024dinov}. This capability is achieved through a knowledge distillation approach, incorporating techniques from DINO~\cite{caron2021emerging} and iBOT~\cite{zhou2022image}, and leveraging a large dataset. The dataset includes both curated collections and additional data sourced from the web. For each image in the curated dataset, a specified number of uncurated nearest neighbors, identified through cosine similarity of the embeddings, are selected to construct an expanded training set.
Despite being unsupervised, DINOv2 achieves performance comparable to state of the art weakly supervised models. Its effectiveness spans various downstream tasks, such as classification, semantic segmentation, and depth estimation, demonstrating its versatility across domains.

We employ the DINOv2 model to both predict the salient patches as visual fixations (Fig.~\ref{fig:overview}a and Section~\ref{sec:salient}) and to extract the features with the pre-trained weights for the classification task as the evaluation of the retinal prosthetic simulation and the corresponding optimization, as shown in Fig.~\ref{fig:overview}d. 
We argue that the pre-trained, frozen DINOv2 backbone offers an unsubstantiated simulation of the innate human ability to recognize visual patterns, which can be further validated through subjective assessments with patients. To evaluate the model’s efficacy across various approaches, whether based on downsampling or fixation and whether encoded or not, we utilize linear probing to measure how effectively the model’s representations transfer to the classification task. High accuracy in linear probing suggests that the stimulation process yields robust and meaningful phosphenes. Additionally, the improvement in classification with learnable linear probing can be viewed as analogous to the learning process during rehabilitation following the implantation of retinal prosthetics.

\section{Experiments and Results}

\subsection{Simulation of Fixation-based Healthy Vision}

Inspired by the saccade mechanism of the human eye \cite{martinez2004role}, healthy vision is simulated by selecting a limited number of salient fixation patches from an input still image of the Imagenette dataset without a percept simulator. 
The performance is evaluated through classification accuracy using the pre-trained DINOv2 ViT-S/14 distilled without registers \cite{oquab2024dinov}, chosen for its efficiency as a lightweight model. 
Neither the encoder nor the simulator has been implemented in this approach. 
We report an upper-bound accuracy of 92.76\% using only 10\% of the salient patches (Section \ref{sec:salient}), as listed in Table~\ref{tab:result}. The preserved fixations are shown in Fig.~\ref{fig:visualization}b.

\subsection{Simulation of Prosthetic Vision}

In the following experiments, we use the physiologically validated Axon Map Model \cite{beyeler2019model} to predict visual percepts from electrical stimuli without applying an encoder (Fig. \ref{fig:overview}b).

\begin{figure}[t!]
    \centering
    \includegraphics[width=.95\linewidth]{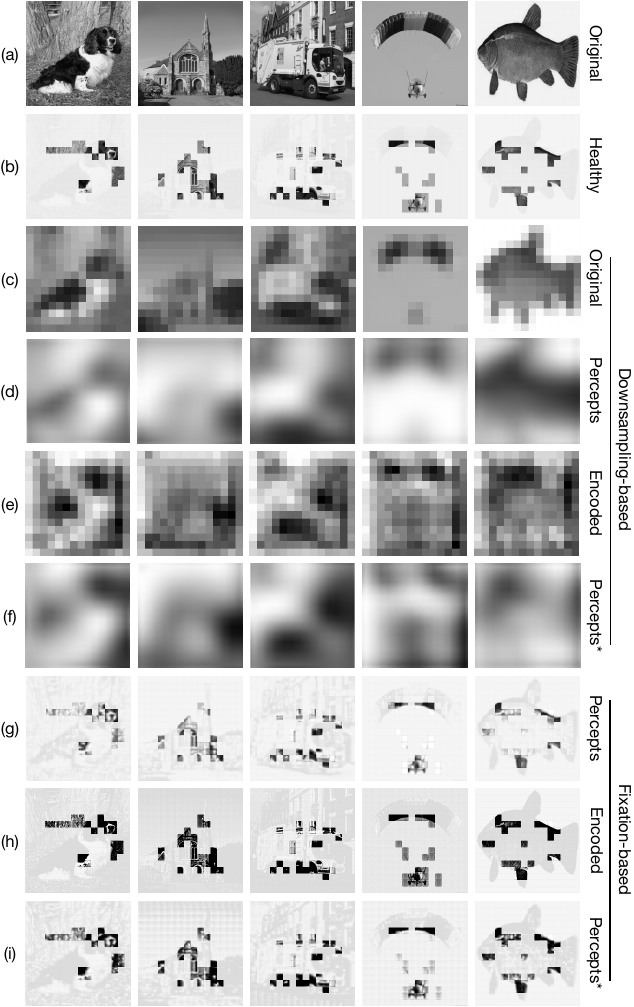}
    \caption{Visualization. Simulated visual fixations are preserved in (b) and (g-i), best viewed when zoomed in. Percepts refer to the predicted phosphenes generated from the original stimuli, while percepts* are those from the encoded stimuli with a U-Net \cite{ronneberger2015u}. The Axon Map Model \cite{beyeler2019model} is used to predict percepts with realistic parameters $\rho$ and $\lambda$.}
    \label{fig:visualization}
\end{figure}

\noindent\textbf{Downsampling-based.}
Due to the limited resolution of the electrode array in the retinal implant, the input image must be downsampled, leading to a loss of information. Examples of images downsampled from $224\times 224$ to $14\times 14$ pixels are presented in Fig~\ref{fig:visualization}c. The corresponding percepts, predicted using the Axon Map Model, are illustrated in Fig.~\ref{fig:visualization}d. Using the frozen DINOv2 model, we achieve a baseline classification accuracy of 47.46\% under idealized conditions and 38.70\% under more realistic conditions (Table~\ref{tab:result}).

\noindent\textbf{Fixation-based.} 
Rather than downsampling the entire image, we propose using only 10\% of salient patches (Section~\ref{sec:salient}) as the input stimulus for the electrode array. By mimicking the saccade mechanism, we can alleviate the limitations posed by the resolution of the electrode array. As demonstrated in Fig.~\ref{fig:visualization}g, the percepts tend to appear brighter (less distinguishable with higher intensity) compared to the healthy case (Fig.~\ref{fig:visualization}b), due to heavy radial and axonal distortion (Section~\ref{sec:simulator}). Nevertheless, as shown in Table~\ref{tab:result}, the fixation-based stimulus produces more semantically understandable percepts than the downsampling-based approach, resulting in a classification accuracy improvement from 47.46\% to 85.20\% with idealized parameters and from 38.70\% to 81.99\% with realistic parameters.

\subsection{Optimization of Prosthetic Vision}

As introduced in Section~\ref{sec:encoder}, we train an encoder in an end-to-end manner with a U-Net architecture \cite{ronneberger2015u}.

\noindent\textbf{Downsampling-based.}
Although the classification accuracy improves with the U-Net encoder (by 4\% with idealized and 2\% with realistic $\rho$ and $\lambda$, respectively), the encoded stimulus tends to make aggressive changes (Fig.~\ref{fig:visualization}c \& e), and the optimized percepts may become even less recognizable (Fig.~\ref{fig:visualization}f).

\noindent\textbf{Fixation-based.}
Training an encoder is not always stable, as seen in Table~\ref{tab:result} with either a linear layer or a randomly initialized U-Net (denoted as U-Net*). Pre-training the network to generate outputs identical to the input has been found to improve stability of subsequent end-to-end training simultaneously with the Axon Map Model \cite{beyeler2019model} and the DINOv2 \cite{oquab2024dinov}. During identity weight initialization, the U-Net learns to replicate the random input using the mean squared error loss. The contrast of the optimized percepts (Fig.~\ref{fig:visualization}i) given the encoded stimuli (Fig.~\ref{fig:visualization}h) is higher compared to the non-optimized ones (Fig.~\ref{fig:visualization}g), yielding classification accuracy near the upper bound: 90.85\% with idealized parameters and 87.72\% with realistic parameters.

\begin{table}[t]
    \caption{Comparison of classification accuracy (in \%) for different approaches on the validation set. U-Net refers to identity weight initialization, while U-Net* indicates random weight initialization. The Axon Map Model is used with parameters $\rho=150\,\mu m, \lambda=100\,\mu m$ for Param. A, and $\rho=437\,\mu m, \lambda=1420\,\mu m$ for Param. B, respectively.}
    \centering
    \small
    \begin{tabular}{ccccc}
    \toprule
      \multirow{2}{3em}[-0.2em]{Fixation}    & \multirow{2}{3em}[-0.2em]{Encoder}  & \multirow{2}{6em}[-0.2em]{Linear probing}  & \multicolumn{2}{c}{Accuracy (in \%)}\\\cmidrule{4-5}
      & & & Param. A  & Param. B \\\midrule
    \ding{55} & \ding{55} & Learnable & 47.46 & 38.70 \\
        \ding{55} & U-Net & Learnable & 51.49 & 40.59 \\\midrule
        \ding{51} & \ding{55} & Learnable & 85.20 & 81.99 \\
       \ding{51}  & Linear & Learnable & 21.12 & 21.43 \\
    \ding{51}     & U-Net* & Learnable & 53.81 & 68.79 \\\midrule
    \ding{51} & U-Net & Frozen & 85.94 & 82.62 \\
       \ding{51}  & U-Net &  Learnable &  \textbf{90.85} & \textbf{87.72} \\\midrule
       \multicolumn{3}{c}{Healthy upper bound as comparate:} & \multicolumn{2}{c}{92.76}\\\bottomrule
    \end{tabular}
    \label{tab:result}
\end{table}

\section{Conclusion}

We present a novel retinal prosthetic \textit{in silico} framework that leverages visual fixations inspired by the saccade mechanism of the human eye as an alternative to downsampling, enabling a potentially more realistic and efficient simulation of retinal prosthetic vision. By employing end-to-end optimization with an encoder, a physiologically validated visual percept simulator, and a pre-trained foundation model, we achieve improved classification accuracy, marking a promising advancement in retinal prosthetics. Future work will explore incorporating temporal dynamics of fixation points and refining the encoder to prioritize human recognizability, mitigating its reliance on network-specific features only for classification.

\section{Compliance with Ethical Standards}
Ethical approval was not required as the study relied solely on open-source datasets and computational models.

\section{Acknowledgements}
This work was supported by Deutsche Forschungsgemeinschaft (DFG, German Research Foundation) with the grant GRK2610: InnoRetVision (project number 424556709).

\bibliographystyle{IEEEbib}
\small
\bibliography{bib}

\begin{thebibliography}{10}

\bibitem{Ayton_2020}
Lauren~N. Ayton, Nick Barnes, Gislin Dagnelie, Takashi Fujikado, Georges Goetz, Ralf Hornig, Bryan~W. Jones, Mahiul~M.K. Muqit, Daniel~L. Rathbun, Katarina Stingl, James~D. Weiland, and Matthew~A. Petoe,
\newblock ``An update on retinal prostheses,''
\newblock {\em Clinical Neurophysiology}, vol. 131, no. 6, pp. 1383--1398, 2020.

\bibitem{michael_beyeler-proc-scipy-2017}
Michael Beyeler, Geoffrey~M. Boynton, Ione Fine, and Ariel Rokem,
\newblock ``pulse2percept: A python-based simulation framework for bionic vision,''
\newblock in {\em Proceedings of the 16th Python in Science Conference}, 2017, pp. 81--88.

\bibitem{luo2016argus}
Yvonne Hsu-Lin Luo and Lyndon Da~Cruz,
\newblock ``The {Argus}{\textregistered} {II} retinal prosthesis system,''
\newblock {\em Progress in Retinal and Eye Research}, vol. 50, pp. 89--107, 2016.

\bibitem{Granley2022hybrid}
Jacob Granley, Lucas Relic, and Michael Beyeler,
\newblock ``Hybrid neural autoencoders for stimulus encoding in visual and other sensory neuroprostheses,''
\newblock in {\em Advances in Neural Information Processing Systems}, 2022, vol.~35, pp. 22671--22685.

\bibitem{Relic2022deep}
Lucas Relic, Bowen Zhang, Yi-Lin Tuan, and Michael Beyeler,
\newblock ``Deep learning–based perceptual stimulus encoder for bionic vision,''
\newblock in {\em Proceedings of the Augmented Humans International Conference}, 2022, pp. 323--325.

\bibitem{RuytervanSteveninck2022}
Jaap de~Ruyter~van Steveninck, Umut Güçlü, Richard van Wezel, and Marcel van Gerven,
\newblock ``End-to-end optimization of prosthetic vision,''
\newblock {\em Journal of Vision}, vol. 22, no. 2, pp. 20, 2022.

\bibitem{wu2023embc}
Yuli Wu, Ivan Karetic, Johannes Stegmaier, Peter Walter, and Dorit Merhof,
\newblock ``A deep learning-based in silico framework for optimization on retinal prosthetic stimulation,''
\newblock in {\em International Conference of the IEEE Engineering in Medicine \& Biology Society (EMBC)}, 2023, pp. 1--4.

\bibitem{Granley2023human}
Jacob Granley, Tristan Fauvel, Matthew Chalk, and Michael Beyeler,
\newblock ``Human-in-the-loop optimization for deep stimulus encoding in visual prostheses,''
\newblock in {\em Advances in Neural Information Processing Systems}, 2023, vol.~36, pp. 79376--79398.

\bibitem{wu2024optimizing}
Yuli Wu, Julian Wittmann, Peter Walter, and Johannes Stegmaier,
\newblock ``Optimizing retinal prosthetic stimuli with conditional invertible neural networks,''
\newblock {\em arXiv preprint arXiv:2403.04884}, 2024.

\bibitem{EricksonDavis2021}
Cordelia Erickson-Davis and Helma Korzybska,
\newblock ``What do blind people “see” with retinal prostheses? {O}bservations and qualitative reports of epiretinal implant users,''
\newblock {\em PLOS ONE}, vol. 16, no. 2, pp. 1--23, 2021.

\bibitem{martinez2004role}
Susana Martinez-Conde, Stephen~L. Macknik, and David~H. Hubel,
\newblock ``The role of fixational eye movements in visual perception,''
\newblock {\em Nature Reviews Neuroscience}, vol. 5, no. 3, pp. 229--240, 2004.

\bibitem{Kuemmerer2022}
Matthias Kümmerer, Matthias Bethge, and Thomas S.~A. Wallis,
\newblock ``{DeepGaze III}: Modeling free-viewing human scanpaths with deep learning,''
\newblock {\em Journal of Vision}, vol. 22, no. 5, pp. 7, 2022.

\bibitem{cornia2018predicting}
Marcella Cornia, Lorenzo Baraldi, Giuseppe Serra, and Rita Cucchiara,
\newblock ``Predicting human eye fixations via an {LSTM}-based saliency attentive model,''
\newblock {\em IEEE Transactions on Image Processing}, vol. 27, no. 10, pp. 5142--5154, 2018.

\bibitem{wiki:saccade}
Wikimedia Commons,
\newblock ``File:szakkad.jpg --- wikimedia commons{,} the free media repository,'' 2022,
\newblock [Online; accessed 18-September-2024].

\bibitem{oquab2024dinov}
Maxime Oquab, Timothée Darcet, Théo Moutakanni, Huy~V. Vo, Marc Szafraniec, Vasil Khalidov, Pierre Fernandez, Daniel Haziza, Francisco Massa, Alaaeldin El-Nouby, Mahmoud Assran, Nicolas Ballas, Wojciech Galuba, Russell Howes, Po-Yao Huang, Shang-Wen Li, Ishan Misra, Michael Rabbat, Vasu Sharma, Gabriel Synnaeve, Hu~Xu, Hervé Jegou, Julien Mairal, Patrick Labatut, Armand Joulin, and Piotr Bojanowski,
\newblock ``{DINO}v2: Learning robust visual features without supervision,''
\newblock {\em Transactions on Machine Learning Research}, 2024.

\bibitem{dosovitskiy2021an}
Alexey Dosovitskiy, Lucas Beyer, Alexander Kolesnikov, Dirk Weissenborn, Xiaohua Zhai, Thomas Unterthiner, Mostafa Dehghani, Matthias Minderer, Georg Heigold, Sylvain Gelly, Jakob Uszkoreit, and Neil Houlsby,
\newblock ``An image is worth 16x16 words: Transformers for image recognition at scale,''
\newblock in {\em International Conference on Learning Representations}, 2021.

\bibitem{caron2021emerging}
Mathilde Caron, Hugo Touvron, Ishan Misra, Herv{\'e} J{\'e}gou, Julien Mairal, Piotr Bojanowski, and Armand Joulin,
\newblock ``Emerging properties in self-supervised vision transformers,''
\newblock in {\em Proceedings of the IEEE/CVF International Conference on Computer Vision}, 2021, pp. 9650--9660.

\bibitem{ronneberger2015u}
Olaf Ronneberger, Philipp Fischer, and Thomas Brox,
\newblock ``{U-Net}: Convolutional networks for biomedical image segmentation,''
\newblock in {\em International Conference on Medical Image Computing and Computer-Assisted Intervention}, 2015, pp. 234--241.

\bibitem{deng2009imagenet}
Jia Deng, Wei Dong, Richard Socher, Li-Jia Li, Kai Li, and Li~Fei-Fei,
\newblock ``{ImageNet}: A large-scale hierarchical image database,''
\newblock in {\em IEEE Conference on Computer Vision and Pattern Recognition}, 2009, pp. 248--255.

\bibitem{beyeler2019model}
Michael Beyeler, Devyani Nanduri, James~D. Weiland, Ariel Rokem, Geoffrey~M. Boynton, and Ione Fine,
\newblock ``A model of ganglion axon pathways accounts for percepts elicited by retinal implants,''
\newblock {\em Scientific Reports}, vol. 9, no. 1, pp. 1--16, 2019.

\bibitem{vaswani2017attention}
Ashish Vaswani, Noam Shazeer, Niki Parmar, Jakob Uszkoreit, Llion Jones, Aidan~N. Gomez, \L{}ukasz Kaiser, and Illia Polosukhin,
\newblock ``Attention is all you need,''
\newblock in {\em Proceedings of the 31st International Conference on Neural Information Processing Systems}, 2017, pp. 6000--6010.

\bibitem{he2022masked}
Kaiming He, Xinlei Chen, Saining Xie, Yanghao Li, Piotr Doll{\'a}r, and Ross Girshick,
\newblock ``Masked autoencoders are scalable vision learners,''
\newblock in {\em Proceedings of the IEEE/CVF Conference on Computer Vision and Pattern Recognition}, 2022, pp. 16000--16009.

\bibitem{greenberg1999computational}
Robert~J. Greenberg, Toby~J. Velte, Mark~S. Humayun, George~N. Scarlatis, and Eugene de~Juan~Jr.,
\newblock ``A computational model of electrical stimulation of the retinal ganglion cell,''
\newblock {\em IEEE Transactions on Biomedical Engineering}, vol. 46, no. 5, pp. 505--514, 1999.

\bibitem{zhou2022image}
Jinghao Zhou, Chen Wei, Huiyu Wang, Wei Shen, Cihang Xie, Alan Yuille, and Tao Kong,
\newblock ``Image {BERT} pre-training with online tokenizer,''
\newblock in {\em International Conference on Learning Representations}, 2022.

\end{thebibliography}

\end{document}